\documentclass{article}

\usepackage[final]{neurips_2024}

\usepackage{amsmath}
\usepackage{booktabs}
\usepackage{multirow}
\usepackage{makecell}
\usepackage{booktabs}

\usepackage[utf8]{inputenc} 
\usepackage[T1]{fontenc}    
\usepackage{hyperref}       
\usepackage{url}            
\usepackage{booktabs}       
\usepackage{amsfonts}       
\usepackage{nicefrac}       
\usepackage{microtype}      
\usepackage{xcolor}         
\usepackage{graphicx}
\usepackage[multiple]{footmisc}
\usepackage{float}
\usepackage{caption}
\usepackage{subcaption}

\title{Improving Multi-candidate Speculative Decoding}

\let\SUP\textsuperscript
\author{
  Xiaofan Lu\thanks{Contributed equally} \SUP{ , }\SUP{1}, Yixiao Zeng\footnotemark[1] \SUP{ , }\SUP{2}, Feiyang Ma\SUP{3}, Zixu Yu\SUP{4}, Marco Levorato\SUP{5}\\
  University of California, Irvine \\
  \texttt{\{\SUP{1}xiaofl14, \SUP{2}yixiaz8, \SUP{3}feiyangm, \SUP{4}zixuy, \SUP{5}levorato\}@uci.edu}
}

\begin{document}
\maketitle
\begin{abstract}
Speculative Decoding (SD) is a technique to accelerate the inference of Large Language Models (LLMs) by using a lower complexity draft model to propose candidate tokens verified by a larger target model. To further improve efficiency, Multi-Candidate Speculative Decoding (MCSD) improves upon this by sampling multiple candidate tokens from the draft model at each step and verifying them in parallel, thus increasing the chances of accepting a token and reducing generation time. Existing MCSD methods rely on the draft model to initialize the multi-candidate sequences and use static length and tree attention structure for draft generation. However, such an approach suffers from the draft and target model's output distribution differences, especially in a dynamic generation context. In this work, we introduce a new version of MCSD that includes a target model initialized multi-candidate generation, a dynamic sliced topology-aware causal mask for dynamic length adjustment, and decision models to optimize early stopping. We experimented with our method on Llama 2-7B and its variants and observed a maximum \textbf{27.5\%} speedup compared to our MCSD baseline across three benchmarks with Llama 2-7B as the target model and JackFram 68M as the draft model. Additionally, we evaluate the effects of using the target model initialized multi-candidate process with different draft models on output quality. Our original code is available on \href{https://github.com/JackZeng0208/DynaSD}{GitHub}.
\end{abstract}
\section{Introduction}
In recent years, Large Language Models (LLMs) such as GPT-4\cite{achiam2023gpt} have significantly advanced various language processing tasks. However, these models are computationally intensive especially during the inference phase, where generating $k$ tokens requires $k$ serial runs of the model. This inefficiency limits the practical deployment of these powerful models in real-time applications.\par
Among serious LLM inference optimization methods, Speculative Decoding (SD)\cite{leviathan2023fast} has been shown to increase inference speed with a marginal generation quality loss. SD frameworks first generate candidate tokens using a model (the draft model) with lower complexity compared to the original model (the target model). Then, the target model verifies the generated tokens. The performance of SD is mainly determined by the token acceptance rate $\alpha$, which measures the proportion of candidate tokens generated by the draft model that the target model accepts. The benefits of using a faster draft model diminish if the target model frequently rejects these tokens, which means that the target model must re-generate the tokens.\par

To further enhance the acceptance rate, Multi-Candidate Speculative Decoding (MCSD)\cite{10.1145/3620666.3651335,yang2024multi} was introduced. MCSD samples multiple candidate tokens at each generation step and verifies them in parallel using the target model. This approach increases the likelihood that at least one of the candidate tokens will be accepted, thereby improving the overall acceptance rate and efficiency. MCSD also incorporates a tree attention mechanism to manage computational and communication overhead by organizing multiple candidate sequences into a single sequence and applying a carefully designed attention mask. However, MCSD still faces several challenges. \textbf{1. Increasing Computational Complexity}: Verifying multiple candidates simultaneously increases the computational load, requiring more memory and processing power. \textbf{2. Efficient Topology-Aware Causal Mask Generation}: Generating and maintaining a topology-aware causal mask for multi-candidate token trees is time-consuming and reduces the adaptivity of the model. \textbf{3. Fixed Draft Generation Length ($\gamma$)}: Using a fixed length for draft-generated token segments may not be optimal in all contexts.\par

In this paper, we present a method that introduces the dynamic sliced topology-aware causal mask to facilitate the speculative decoding process, allowing dynamic adjustment of the draft generation length without reconstructing the topology-aware causal mask. We enhance the acceptance rate by initializing the multi-candidate token tree with the target model, thus improving efficiency. Additionally, we incorporate a decision model to optimize the early stopping mechanism during the draft model generation stage. The model dynamically halts draft token generation early by predicting the likelihood of the target model accepting the tokens, thus reducing unnecessary computation.\par

Our experimental results show that our framework, the combination of target-initialized multi-candidate generation, dynamic sliced topology-aware causal mask, and early stop with a decision model, struggles to maintain both generation quality and speedup simultaneously. Instead, we found that our static target model initialized multi-candidate generation alone achieves the highest speedup while preserving the highest generation quality among our experiments. Therefore, we are presenting the speedup results from our static target-initialized multi-candidate generation experiments in section \ref{sec:setup}, and the results of our framework are in the Appendix \ref{idealAccelerateion}.\par

The static target model initialized multi-candidate generation method with the optimal multi-candidate generation configuration we find through grid search on custom dataset improves generation speed through the improvement in the acceptance rate ($\alpha$), defined as the ratio of the longest draft sequence length accepted by the target model to the maximum draft sequence length. The method achieves a maximum of \textbf{27.5\%} in generation speedup comparing with MCSD baseline and using smaller draft model (Llama-68M \cite{10.1145/3620666.3651335}) on three datasets:  TriviaQA \cite{joshi-etal-2017-triviaqa}, Alpaca \cite{alpaca} and MT-Bench \cite{zheng2024judging}. Output quality evaluation on MT-Bench reveals that the target model initialized multi-candidate process does not preserve the target model's output quality; the output quality decreases as the number of target-initialized tokens increases, and different draft models significantly impact output quality. We also conduct an ablation study to evaluate the impact of the decision model in Appendix~\ref{app:decision_model}. Additionally, we analyze why our framework does not outperform static target-initialized multi-candidate generation alone in Session \ref{discussion}. \par


\section{Background}
\subsection{Speculative Decoding}


Speculative decoding is the collaboration of two models: a smaller draft model (often a more efficient approximation model) and a larger target model. First, the draft model generates multiple candidate tokens in parallel, using its probability distribution to predict the following possible tokens based on input. Then, these tokens are passed to the target model, which verifies them by computing their probability distribution over the same input. If the target model accepts the candidate tokens (probabilities alignment), they are finalized in the output sequence. Otherwise, the target model replaces them by generating new tokens based on its own distribution. The speculative decoding process ensures that the output distribution remains consistent with what the target model alone would produce, thus maintaining the quality of the generated content\cite{leviathan2023fast}. \par

Importantly, this technique does not require changes to the model’s architecture or retraining, making it an accessible and efficient solution for accelerating inference.\par
\subsection{Multi-Candidate Generation}
\begin{figure}
    \centering
    \includegraphics[width=\linewidth]{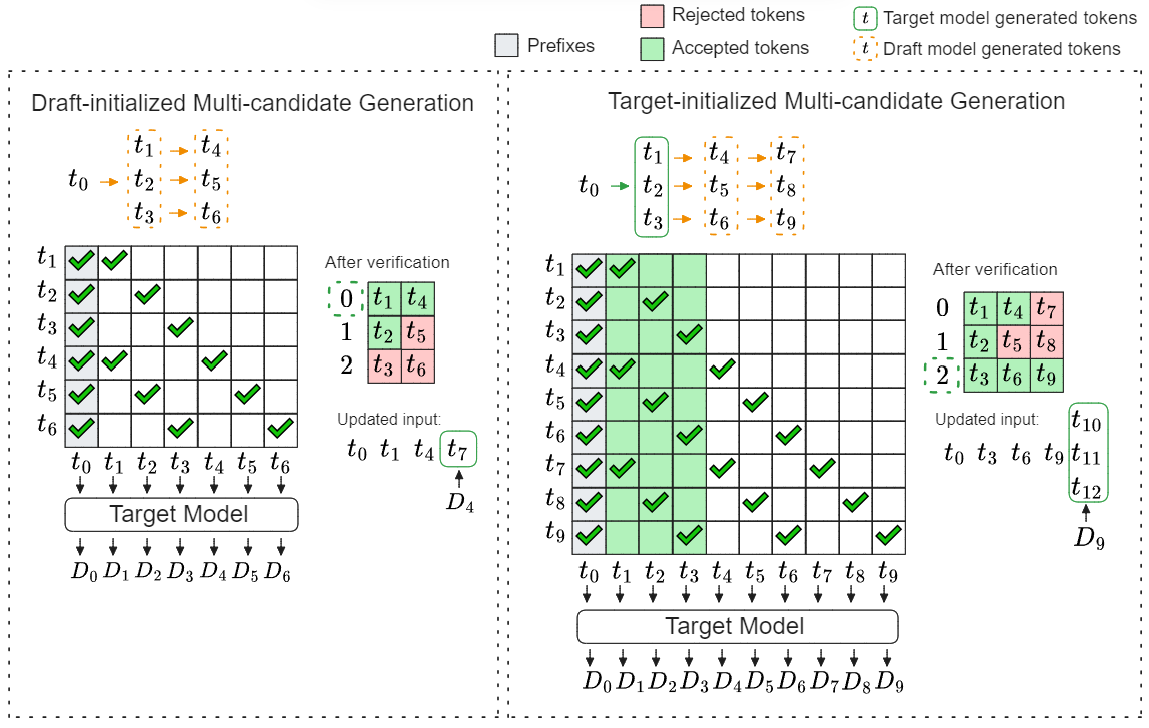}
    \caption{Both the draft-initialized (left) and target-initialized (right) multi-candidate generation processes utilize a token tree configuration with a width of 3 and depth of 2. The execution sequence proceeds as follows: (\textbf{1}) Generate the token tree (shown at the top of each diagram). (\textbf{2}) Transform the token tree into a topology-aware causal mask (represented as a square mask with a check symbol). (\textbf{3}) Generate multi-candidate sequences using the draft model (not shown in the figure).
(\textbf{4}) Verify the multi-candidate sequences with the target model by obtaining next-token logits, which are then transformed into distributions.
(\textbf{5}) Select the candidate sequence with the longest length after verification.
(\textbf{6}) Update the input IDs, key-value cache, and sample new token(s) based on the target model's next-token distributions.
Note: In a draft-initialized multi-candidate generation, only one new token is sampled, whereas in a target-initialized multi-candidate generation, multiple new tokens are sampled.}
    \label{fig:method_compare}
\end{figure}

Due to distributional differences between the draft and target models, the candidate path with the highest probability in the draft model may not always result in the most accepted tokens by the target model. Therefore, verifying multiple candidate paths in parallel increases the overall acceptance rate $\alpha$ of draft tokens. This paper takes the tree attention methodology in Specinfer \cite{10.1145/3620666.3651335} as a starting point to process multiple candidate token paths concurrently. Unlike the traditional causal attention paradigm, which was designed for single text sequence generation, tree attention calculates attention scores for multiple text sequences from a token tree, which requires a topology-aware casual mask plus changes in usual single candidate positional indices and key-value cache update to fuse tree attention computation of all tokens in a single kernel. In Figure \ref{fig:method_compare}, we illustrate the topology-aware causal mask for calculating the tree attention of a multi-candidate sequence with three candidate token paths, each with two draft tokens.
\section{Method}
In the speculative decoding framework proposed by Leviathan et al. \cite{leviathan2023fast}, the expected improvement factor (IF) is defined as:
\begin{equation}
    IF = \frac{1-\alpha^{\gamma+1}}{(1-\alpha)(c\gamma+1)}\nonumber
\end{equation}
where $\alpha$ represents the expectation of acceptance rate, $\gamma$ represent the draft generation length, and $c$ represent the ratio between the time for a single run of draft model and time for a single run of target model. Based on the formula (1), a larger $\alpha$ will speedup token generation. If an oracle could determine $\gamma$ dynamically, the improvement factor can be up to around 60\% larger than the improvement factor with a fixed $\gamma$.\par

In this work, we explore three methods to improve $\alpha$ and determine $\gamma$ dynamically to accelerate the MCSD process:
\begin{enumerate}
    \item Target Model Initialized Multi-Candidate Generation: we introduce a new methodology to construct multi-candidate sequences that improve $\alpha$ over existing approaches;
    \item Dynamic Sliced Topology-Aware Causal Mask: we introduce a method to efficiently create topology-aware casual masks for dynamic multi-candidate generation;
    \item Early Stop Decision Model: we introduce a low-complexity MLP model to determine $\gamma$ dynamically during each draft generation loop. 
    \end{enumerate}

We remark that the above processes can be integrated into a unified framework. The Target Model Initialized Multi-Candidate Token Tree method can be deployed in isolation as it improves $\alpha$ for both static and dynamic MCSD, while $2$ and $3$ are dependent on each other and need the dynamic MCSD.\par

\subsection{Target Model Initialized Multi-Candidate Generation}
Existing multi-candidate speculative decoding methods rely on a draft model to generate the entire multi-candidate token tree, and after verifying the draft generated tokens, only sample \textbf{one token} from the target model or normalized target and draft model's output distribution. Due to the difference between the output distribution of the draft and target model, there is no specific criterion to determine which token sampled from the target model will yield the longest accepted token sequence for future draft model generation. Therefore, we hypothesize that sampling multiple tokens instead of one token from the target model's distribution to initialize a multi-candidate sequence for future draft model generation can increase the acceptance rate. However, When the target model initializes more than one token for multi-candidate generation, selecting an initialized token based on the longest accepted draft token sequence creates a dependency. This means that the acceptance probabilities of the sequential draft tokens influence the probability of accepting the target-sampled token. Consequently, the output distribution no longer aligns with the target model's. Specifically, by selecting the target token $t_i$ that yields the longest accepted draft sequence, the probability of accepting the target sampled token $t_i$ becomes:
\begin{equation}
    P_{\text{output}}(t_i) = P_{\text{target}}(t_i) \times P_{\text{acceptance}}(t_{i+1}, t_{i+2}, \dots \mid t_i) \nonumber
\end{equation}
Where $P_{\text{acceptance}}(t_{i+1}, t_{i+2}, \dots \mid t_i)$ is the probability that the subsequent draft tokens are accepted given the target token $t_i$. This alters the output distribution, causing it to diverge from $P_{\text{target}}(t_i)$. Thus, we evaluate the quality of our method's output using MT-Bench. The empirical results indicate that the quality loss depends on the number of target-driven initialization tokens and the draft model. For more details, please refer to the experimental section.\par

Figure \ref{fig:method_compare} illustrates how we modified the existing draft-initialized multi-candidate generation to build our target model-initialized multi-candidate generation. The topology-aware mask for the target-initialized token tree needs a larger mask than the draft-initialized token tree to handle the multiple initial tokens for different possible sequences sampled from the target model. The topology-aware causal mask size for the target model initialized token tree is the square of the sum of the token tree width and the total number of draft-generated tokens. In Figure \ref{fig:method_compare}, the mask size for the target model initialized token tree is $(3+6)^2$ while the size for the draft initialized token tree is $6^2$. We note that if draft-initialized and target-initialized multi-candidate generation has the same acceptance rate, then the target-initialized method will perform one more target forward pass compared to the draft-initialized method as the origin input passes into the target model instead of the draft model. However, in general, text generation tasks usually involve hundreds of target forward passes, and this overhead is marginal.\par

\begin{figure}
    \centering
    \includegraphics[width=1\linewidth]{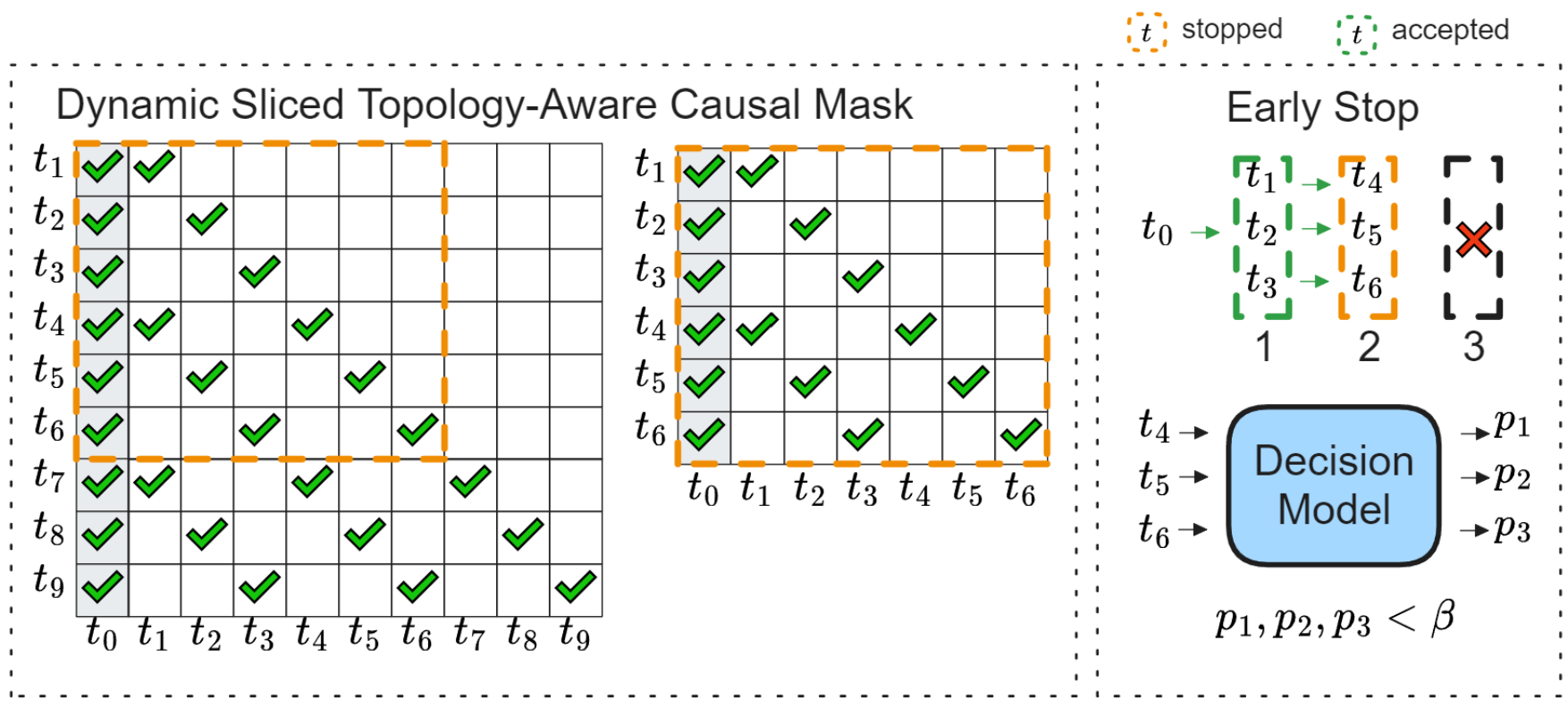}
    \caption{$\beta$ denotes the threshold for early stop, in our experiment the $\beta$ = 0.4. The inputs for the decision model are hidden states or output distribution and entropy related to the token. The dynamic multi-candidate speculative decoding process with an early stop decision model and fork-shaped draft model initialized a token tree, where the token tree configuration is $W = 3$ (width) and $D = 3$ (depth). It stopped at the second draft generation turn, where the maximum number of draft generation turns is three.}
    \label{fig:ds}
\end{figure}

\subsection{Dynamic Sliced Topology-Aware Causal Mask}
Existing multi-candidate speculative decoding methods such as EAGLE~\cite{li2024eagle} employ expansion-based token trees with various depths and widths for different branches to increase the target model's average acceptance rate on draft-generated tokens. Generating a topology-aware casual mask for the multi-candidate token tree is time-consuming; most existing multi-candidate speculative decoding methods only build the topology-aware casual mask once during initialization.\par

We introduce a dynamic sliced topology-aware casual mask to allow the decision model to dynamically decide the length of multi-candidate draft token generation and avoid generating a new topology-aware casual mask during each iteration. The main idea is to construct a large topology-aware casual mask during initialization. Figure \ref{fig:ds} illustrates a topology-aware casual mask that allows a maximum of three draft token generations for each candidate sequence. There are three possible sliced causal masks for dynamic multi-candidate speculative decoding: mask when early stop at first iteration (upper left 3 x 3 causal mask), early stop at second iteration (shown in Figure \ref{fig:ds}), no early stop (entire 9 x 9 casual mask).\par
In the proposed framework, we keep the token tree in fork shape so that each candidate token sequence does not expand to new sub-sequences. The main benefit of using a fork shape instead of an expansion-based token tree is a reduction in memory usage since the size of a topology-aware causal mask for an expansion-based token tree grows exponentially. In contrast, the fork-shaped token tree will grow linearly. \par
\subsection{Decision Model}
We design two types of decision models to dynamically determine whether an early stop is necessary during the draft generation process. The first type of decision model is a three-layer MLP, taking the hidden states from draft model as input and the $max(1,\frac{p(x|I)}{q(x|I)})$ ($p(x|I)$ is the probability of target model predict token $x$ with given input $I$ and $q(x|I)$ is the probability of draft model predict token $x$ with given input $I$) as the training label. The decision process can be represented as:
\begin{equation}
    P_{T1}= MLP(y_i^H)\nonumber
\end{equation}
Where $y_i^H$ denotes the hidden states of the draft model for token $i$. The second type of decision model is inspired by Tandem transformer \cite{s2024tandemtransformersinferenceefficient}, is a two layer-MLP with draft model's output distribution's entropy that takes probabilities as input and the result of verification (zero as rejected, one as accepted) as a label. The decision process can be represented as:
\begin{equation}
    P_{T2} = MLP(ConCat(Topk(y_i^D),Ent(y_i^D))\nonumber
\end{equation}
Where $y_i^D$ represents the output distribution of the draft model for token $i$ and is used to compute both the entropy and top-k probabilities for input to the two layer-MLP. \par

During the draft generation process, the decision model will batch inference the input for multiple sequences and calculate the probabilities of the target model accepting each sequence. If the probabilities for all sequences are lower than the threshold (in our experiment, the threshold is 0.4 for both decision models), then draft generation will stop early; otherwise, it will continue until the maximum draft generation length is reached.\par

\section{Experiment}
\subsection{Experimental Setup}
\label{sec:setup}
Our experiments are performed on a single server equipped with an Nvidia RTX 4090 GPU. To ensure consistency with the original training parameters of the Llama model, all target models use Bfloat16 precision, while draft models are configured with double precision. This setup provides a stable testing environment: Bfloat16 precision on draft models occasionally leads to NaN or Inf values in softmax calculations within the PyTorch library, but double precision for draft models prevents these issues from interrupting the experiment. \par


For our testing dataset, we select TriviaQA \cite{joshi-etal-2017-triviaqa}, Alpaca \cite{alpaca} and MT-Bench \cite{zheng2024judging}. In the following testing, for TriviaQA and Alpaca, we randomly select 250 input prompts. And for MT-bench, we test the complete 80 sets of prompts.\par

For the draft and target models, we extensively utilize the Llama 2-7B Chat model \cite{touvron2023llama} and other related models that share the same tokenizer. As for our target models, we select LLama 2-7B Chat and its fine-tuned version, Vicuna-7B \cite{vicuna2023}. At the same time, we use two different models, the Llama-68M from SpecInfer \cite{10.1145/3620666.3651335} and the TinyLlama-1.1B \cite{zhang2024tinyllama}, as draft models to test the effect of the acceleration methods on different types of model pairs. The maximum generation length is 200 tokens. We apply a temperature of 0 for greedy sampling and 0.7 for probabilistic sampling, with the latter value providing a midpoint that balances the trade-off between generation speed and quality, as observed in our experiments.\par

Furthermore, we standardize the configuration across different SD methods to evaluate the acceptance rate and other performance metrics. Specifically, we set the SD to a fixed $\gamma = 4$, while for the MCSD method, we use the optimal k-configuration of (4,2,2,1) in our environment, as it provides a greater speedup than the (4,2,2,1,1) configuration reported in the original paper \cite{yang2024multi}. \par

Our method employs a static target model initialized MCSD configuration with (2,4,3,1,1), where the first number (2) in the configuration represents the number of target model initialized tokens, and the draft model will generate $4 + 4\times 3 + 4 \times 3 \times 1 +4 \times 3 \times 1\times 1 = 40$  draft tokens for each target model initialized token result $2 \times 4 \times 3\times 1 \times1 = 24$ different candidate sequences with draft token length equal to four ($\gamma = 4$).  Figure \ref{fig:speedup_ratios} shows the speedup ratio under such circumstances. Table \ref{table:quality} shows that our method achieves the highest output quality when the number of target-initialized tokens is set to two. Notably, when the number of target-initialized tokens is reduced to one, the target output distribution remains stable, and the MT-bench score is preserved within a margin of $\pm{0.05}$, regardless of the draft model chosen. Thus, we set our MCSD configuration starting with 2;
through grid search in Figure \ref{fig:speedup_heatmap}, we find (2,4,3,1,1) yields the optimal speedup under our experiment environment on Mt-Bench. Moreover, when the number of target-initialized tokens is equal to two, the greedy sample does not yield a significant improvement in the acceptance rate and ends up with a speedup result similar to the baseline. Therefore, we are not presenting the result with the greedy sample. \par

In addition, we also set dynamic configuration with $D=5$ and $W=16$, where $W$ (Width) indicates the number of candidate sequences generated in parallel at each step of the token tree, and $D$ (Depth) represents the level of token prediction, referring to $\gamma$ in traditional SD. The fixed $D=5$ ensures the draft generation length is consistent across all SD methods. At the same time, $W=16$ aligns the number of candidate sequences with those in MCSD, enabling a fair comparison between the methods. \par
Due to the quality of the generation, all settings in the following test follow the optimal MCSD static configuration. We will illustrate the ideal generation speed and corresponding acceptance rate based on the dynamic configuration in the Appendix section \ref{idealAccelerateion}.\par
\subsection{Overall Results}
\begin{figure}[h!]
    \centering
    \includegraphics[width=0.8\linewidth]{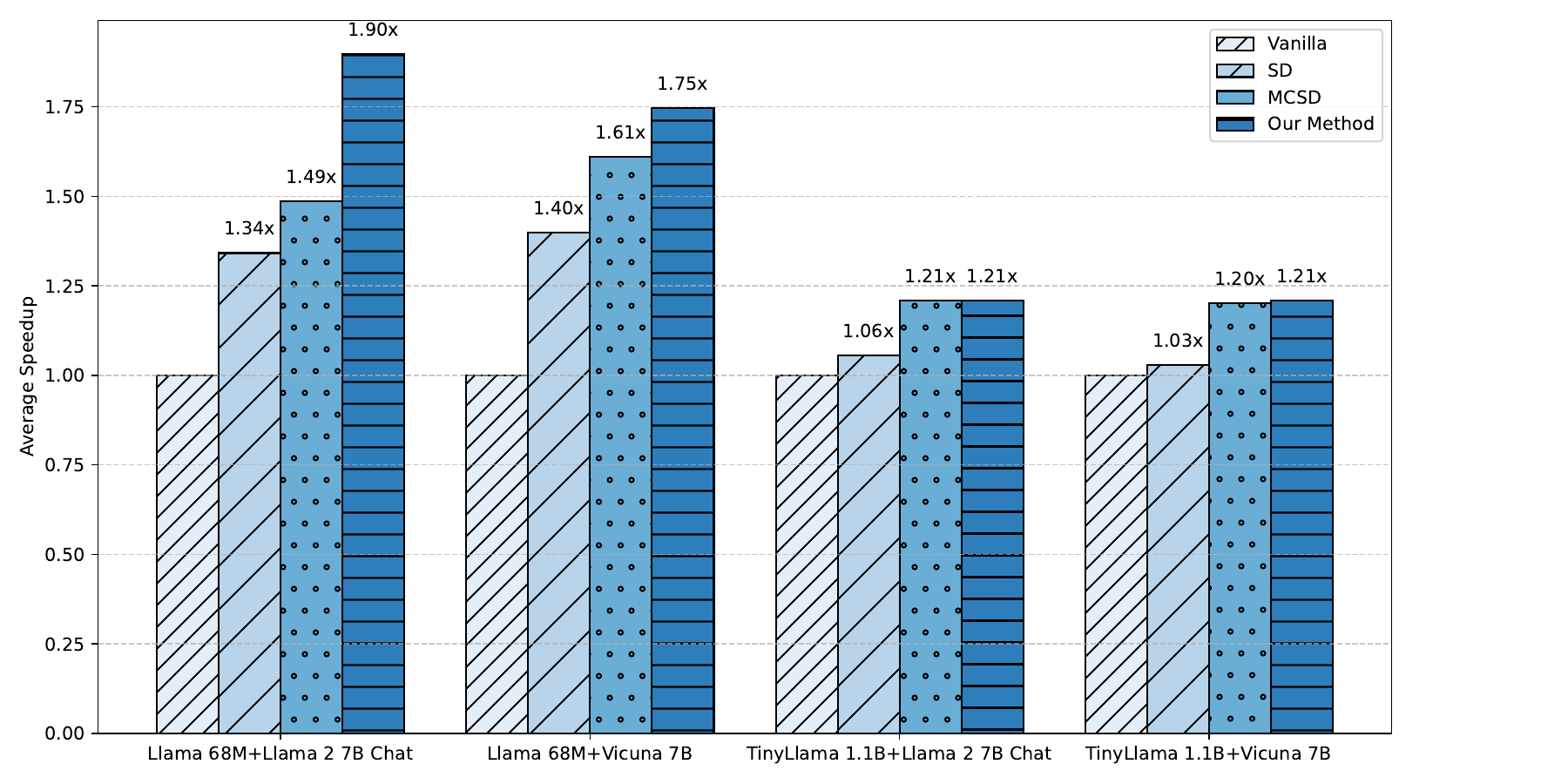}
    \caption{Speedup ratios compared to vanilla inference for different SD methods based on all three datasets under temperature = 0.7. We employ the static tree configuration. The bars represent speedup ratios for different model combinations (draft model + target model)}
    \label{fig:speedup_ratios}
\end{figure}

The experiment result shown in Figure \ref{fig:speedup_ratios} demonstrates substantial improvements in generation speed. All of the configurations are shown and explained in Section \ref{sec:setup}. Conducted on the MT-Bench dataset, our method achieves a maximum speedup of \textbf{1.90} times over the baseline. The primary reason that using LlaMa-68M  as the draft model results in greater speedup compared to TinyLlama-1.1B is that LlaMa-68M  achieves approximately $5.5\times$ faster inference speeds than TinyLlama-1.1B. In contrast, TinyLlama-1.1B shows only a $3.4\times$ higher acceptance rate with Llama-2-7B compared to the acceptance rate of LlaMa-68M.\par

\begin{table}[h!]
    \centering
    \begin{tabular}{ccc}
        \toprule
        \textbf{Target Initialized Width} 
        & \textbf{Llama-68M} 
        & \textbf{TinyLlama-1.1B} \\
        \midrule
        2 & 5.07 (-1.22) & 5.83 (-0.43) \\
        3 & 4.32 (-1.97) & 5.69 (-0.6) \\
        \bottomrule
    \end{tabular}
    \vspace{0.25cm}
    \caption{This shows the impact of target-initialized multi-candidate selection on generation quality (MT-Bench score) across various draft models, with a generation temperature of 0.7. The MT-Bench score (higher the better) for Llama-2-7B at this temperature is 6.29.}
    \label{table:quality}
\end{table}

\begin{table}[h!]
\centering
\begin{tabular}{llccc}
\toprule
\textbf{Dataset} & \textbf{Methods} & \textbf{Configuration} & \textbf{Llama 2-7B Chat} & \textbf{Vicuna-7B} \\
\midrule
\multirow{3}{*}{\textbf{MT-Bench}} 
  & Baseline SD   & $\gamma = 4$                  & 0.17  & 0.18 \\
  & MCSD          & $4\times2\times2\times1$      & 0.29  & 0.31 \\ 
  & Our method    & $2\times4\times3\times1\times1$ & \textbf{0.47}  & \textbf{0.40} \\
\midrule
\multirow{3}{*}{\textbf{TriviaQA}} 
  & Baseline SD   & $\gamma = 4$                  & 0.15  & 0.21 \\
  & MCSD          & $4\times2\times2\times1$      & 0.21  & 0.33 \\ 
  & Our method    & $2\times4\times3\times1\times1$ & \textbf{0.53}  & \textbf{0.51} \\
\midrule
\multirow{3}{*}{\textbf{Alpaca}} 
  & Baseline SD   & $\gamma = 4$                  & 0.20  & 0.21 \\
  & MCSD          & $4\times2\times2\times1$      & 0.34  & 0.35 \\ 
  & Our method    & $2\times4\times3\times1\times1$ & \textbf{0.52}  & \textbf{0.46} \\
\bottomrule
\end{tabular}
\vspace{0.25cm}
\caption{Comparison of acceptance rate ($\alpha$) for different methods using Llama-68M  as draft model with temperature = 0.7 under MCSD static configuration}
\label{table:acc_rate_comparison}
\end{table}

\section{Discussion}
\label{discussion}
We acknowledge that the combination of target-initialized multi-candidate generation, dynamic sliced topology-aware causal masking, and early stopping with a decision tree does not outperform static target model-initialized multi-candidate generation alone in terms of inference speed while preserving the highest generation quality. Two main factors contribute to this outcome:
\begin{enumerate}
    \item In the dynamic MCSD generation, we employ a fork-shaped token tree. Since the highest generation quality is maintained when the number of target model-initialized tokens is set to two, the fork-shaped token tree produces only two candidate sequences. This results in a lower acceptance rate compared to the tree-shaped token tree used in static target model-initialized multi-candidate generation.
    \item The decision models we trained do not yield a substantial speedup via dynamic $\gamma$ (shown in Appendix \ref{app:decision_model}) . Experimental results indicate that although early stopping provided by the decision model reduces draft generation time by avoiding likely rejected sequences, it also prematurely stops sequences that could validly extend for a longer acceptance length, increasing target model inference time. Consequently, the combined overhead from the decision model inference time and the extended target model inference time offsets the time saved in draft generation, leading to negligible speedup in most cases. Figure \ref{fig:hard_label} suggests why our decision model leads to premature stopping.
\end{enumerate} 
Although we could address the first issue by sampling multiple draft tokens for each target model-initialized token and constructing the fork-shaped token tree based on these expanded draft tokens, this adjustment would not yield significant benefits without a decision model capable of accelerating MCSD inference. \par

We hope our work offers insights for future improvements in the speculative decoding process. For instance, while the dynamic MCSD process could theoretically enhance inference speed, our experimental results suggest that training an external decision model to perform early stopping—even if it could perfectly avoid premature stops and introduce no additional overhead to target model inference—would result in an overall speedup of no more than $10\%$ compared to an optimal static MCSD process. Furthermore, MT-Bench results for target model-initialized multi-candidate generation suggest that when more than one target model-initialized token is used, a draft model with a higher acceptance rate relative to the target model preserves better target output quality than one with a lower acceptance rate. For instance, in a basic speculative decoding setup, TinyLlama-1.1B achieves an acceptance rate approximately $ 25\% $ higher with Llama-2-7B compared to Llama-68M.


\section{Conclusions}

In this work, we propose a target-initialized multi-candidate token tree that enhances the acceptance rate in multi-candidate speculative decoding, with the output quality loss influenced by the number of target-initialized tokens and the specific draft model used. Additionally, we introduced a dynamic mask-slicing technique to avoid topology-aware causal mask generation overhead for dynamic multi-candidate speculative decoding. While we have yet to discover the decision model that could make dynamic multi-candidate speculative decoding faster than static one under all scenarios, our work can help future research find a more efficient multi-candidate speculative decoding process.

\clearpage
\appendix
\section{Appendix}
\begin{figure}[h!]
    \centering
    \includegraphics[width=0.7\linewidth]{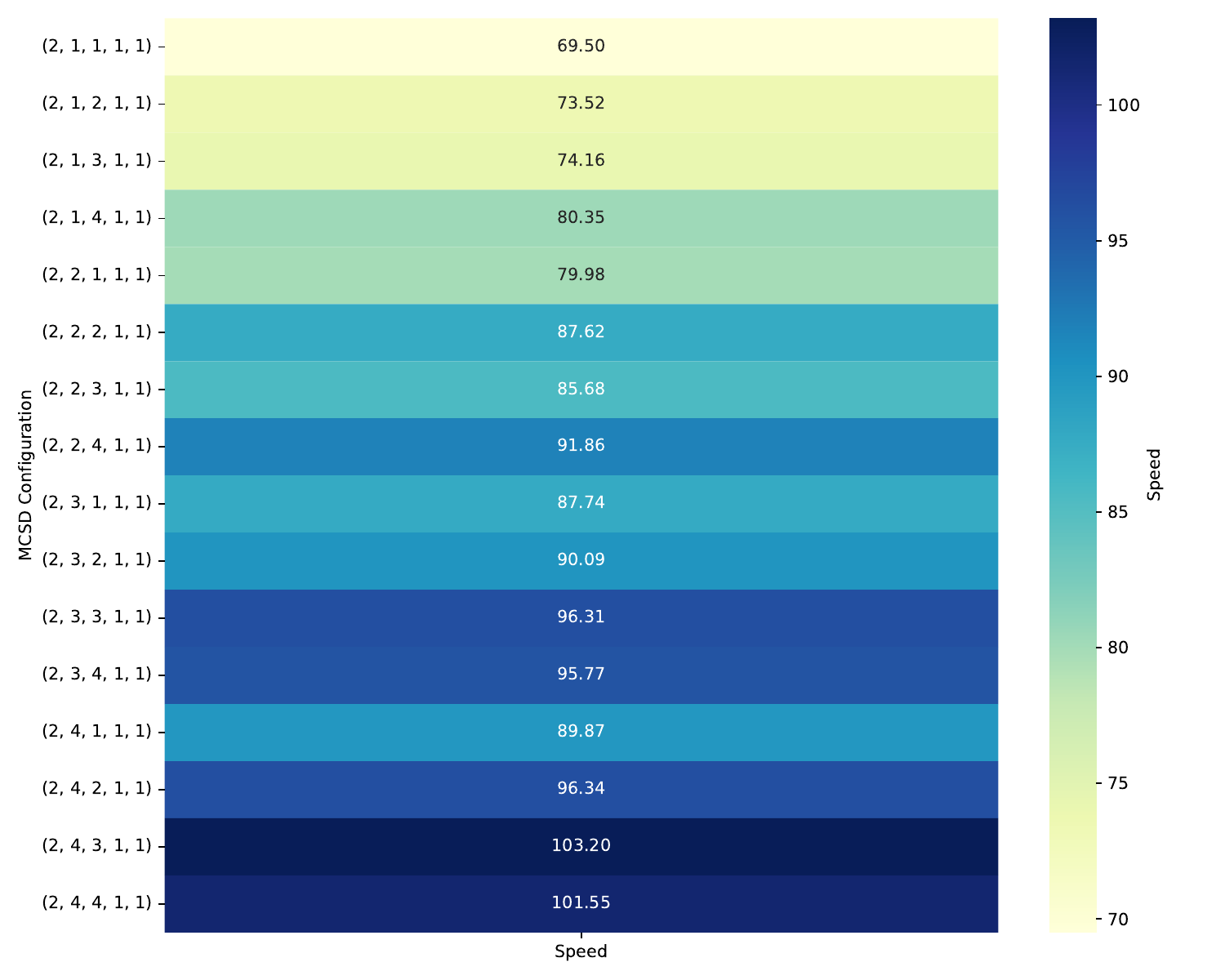}
    \caption{Speed heatmap across various static MCSD configuration}
    \label{fig:speedup_heatmap}
\end{figure}
\subsection{Dynamic MCSD Configurations and Results}
This section thoroughly explores the results and configurations tested in our experiments with dynamic MCSD configuration. This section delves into the impact of various parameter settings on performance metrics, such as speed and acceptance rate, across different model configurations. Specifically, we analyze the effects of dynamic depth and width adjustments and acceptance rates achieved under different draft and target model combinations. Through these insights, this section aims to demonstrate our optimal outcomes when ignoring the quality lost. \par
\label{idealAccelerateion}
\begin{figure}[h!]
    \centering
    \includegraphics[width=0.8\linewidth]{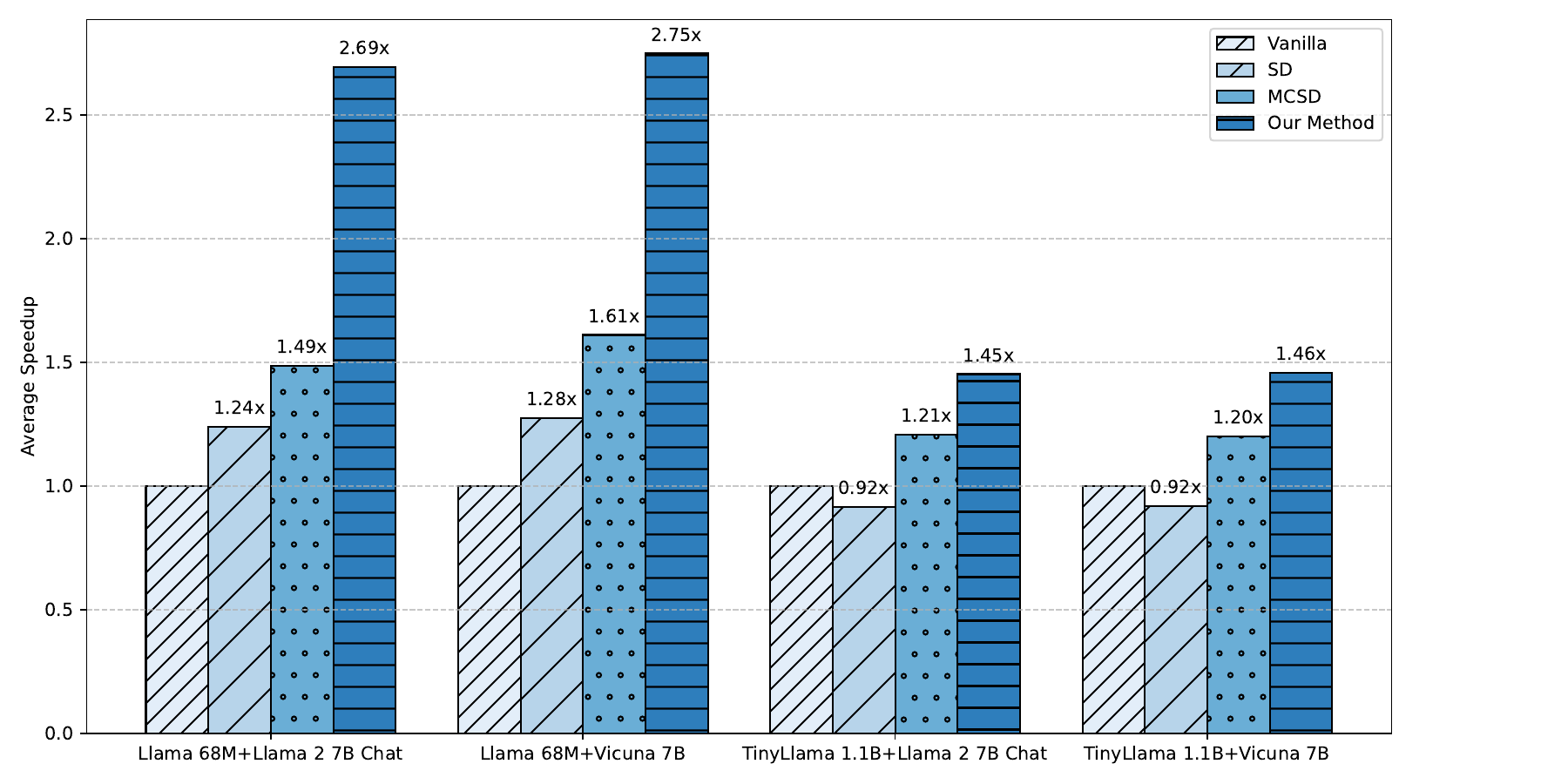}
    \caption{Speedup ratios compared to vanilla inference for different SD methods based on all three datasets under dynamic setting and temperature = 1. We employ dynamic tree configuration with $D = 5$ and $W = 16$. This is our optimal configuration based on empirical studies.}
    \label{fig:speedup_ratios_dynamic}
\end{figure}

\begin{table}[h!]
\centering
\begin{tabular}{llccc}
\toprule
\textbf{Dataset} & \textbf{Methods} & \textbf{Configuration} & \textbf{Llama 2-7B Chat} & \textbf{Vicuna-7B} \\
\midrule
\multirow{3}{*}{\textbf{MT-Bench}} 
  & Baseline SD   & $\gamma = 5$                    & 0.11  & 0.13 \\
  & MCSD          & $4\times2\times2\times1\times1$ & 0.20  & 0.23 \\ 
  & Our method    & $D = 5, W = 16$                 & \textbf{0.74}  & \textbf{0.75} \\
\midrule
\multirow{3}{*}{\textbf{Alpaca}} 
  & Baseline SD   & $\gamma = 5$                    & 0.13  & 0.13 \\
  & MCSD          & $4\times2\times2\times1\times1$ & 0.23  & 0.23 \\ 
  & Our method    & $D = 5, W = 16$                 & \textbf{0.75}  & \textbf{0.79} \\
\midrule
\multirow{3}{*}{\textbf{TriviaQA}} 
  & Baseline SD   & $\gamma = 5$                    & 0.12  & 0.14 \\
  & MCSD          & $4\times2\times2\times1\times1$ & 0.21  & 0.21 \\ 
  & Our method    & $D = 5, W = 16$                 & \textbf{0.75}  & \textbf{0.80} \\
\bottomrule
\end{tabular}
\vspace{0.25cm}
\caption{Comparison of acceptance rate ($\alpha$) for different methods using Llama-68M as draft model with temp = 1. Since $D = 5$, we set $\gamma = 5$ and keep the original optimal setting of MCSD with a maximum tree length of 5}
\label{table:acc_rate_comparison}
\end{table}

\begin{table}[h!]
\centering
\begin{tabular}{llcccc}
\toprule
\multirow{2}{*}{\textbf{Dataset}} & \multirow{2}{*}{\textbf{Draft models}} & \multirow{2}{*}{\textbf{Temp}} & \multicolumn{2}{c}{\textbf{Acceptance Rate $\alpha$}} \\
\cmidrule(lr){4-5}
& & & \textbf{Llama 2-7B Chat} & \textbf{Vicuna-7B} \\
\midrule
\multirow{4}{*}{\textbf{MT-Bench}} 
& Llama-68M & 0 & 0.76 & 0.77 \\
& TinyLlama-1.1B & 0 & 0.93 & 0.93 \\
& Llama-68M & 1 & 0.74 & 0.75 \\
& TinyLlama-1.1B & 1 & 0.95 & 0.95 \\
\midrule
\multirow{4}{*}{\textbf{Alpaca}} 
& Llama-68M      & 0 & 0.80 & 0.83 \\
& TinyLlama-1.1B & 0 & 0.92 & 0.94 \\
& Llama-68M      & 1 & 0.75 & 0.79 \\
& TinyLlama-1.1B & 1 & 0.93 & 0.95 \\
\midrule
\multirow{4}{*}{\textbf{TriviaQA}} 
& Llama-68M & 0 & 0.86 & 0.91 \\
& TinyLlama-1.1B & 0 & 0.96 & 0.94 \\
& Llama-68M & 1 & 0.75 & 0.80 \\
& TinyLlama-1.1B & 1 & 0.93 & 0.94 \\
\bottomrule
\end{tabular}
\vspace{0.25cm}
\caption{Acceptance rates of our methods (temp = 0 and 1, given generation depth $D=5$ and width $W=16$)}
\label{table:dynasd_acc_rate}
\end{table}
\subsubsection{Effects of Width}
\begin{figure}[h]
    \centering
    \includegraphics[width=0.9\linewidth]{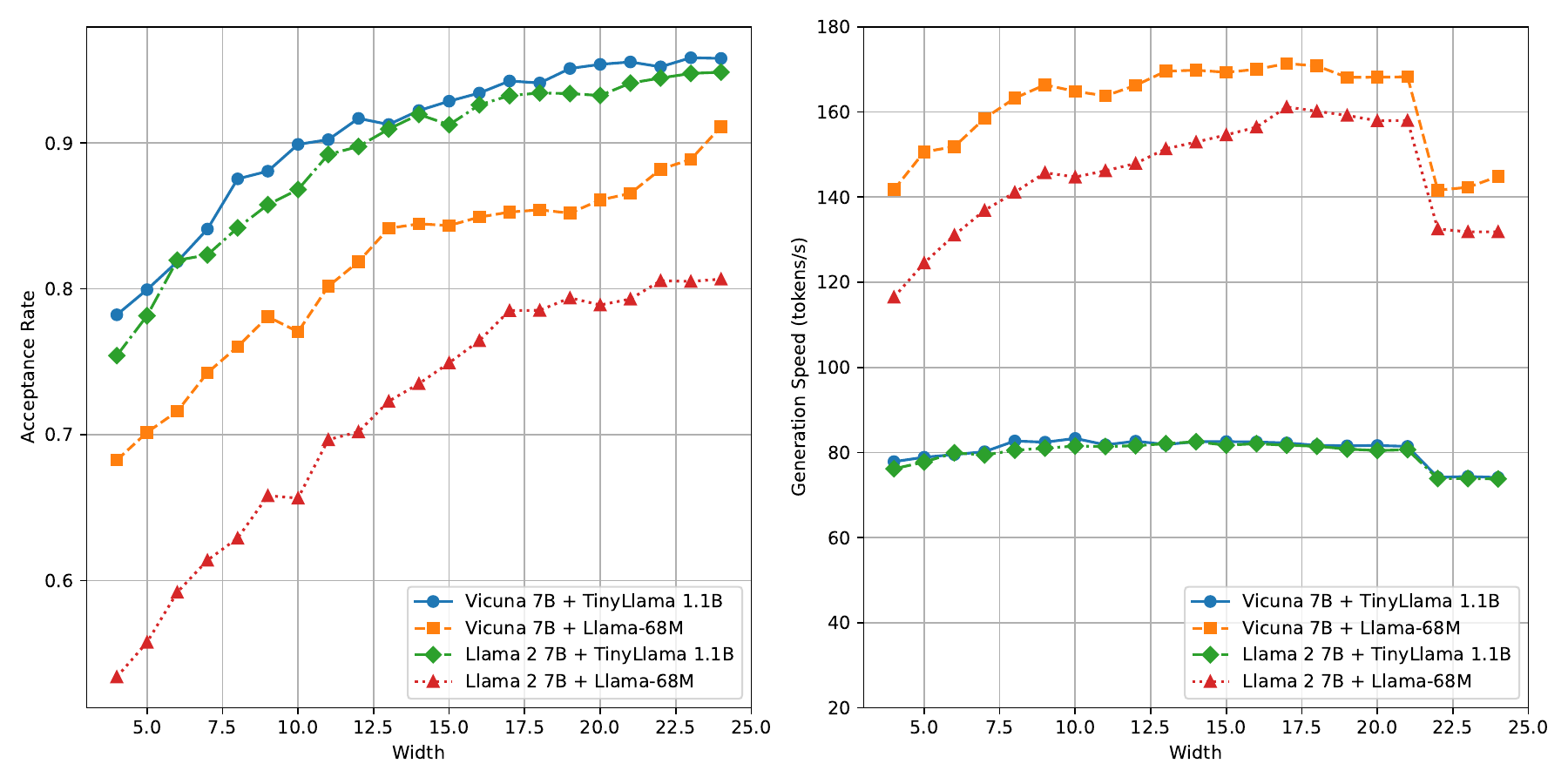}
    \caption{Left graph shows the relationship between width and acceptance rate ($\alpha$) and right graph shows the the relationship between width and generation speed in tokens/s under MCSD dynamic configuration}
    \label{fig:dynasd_width}
\end{figure}
We measure the effect of width on acceptance rate and speed in Figure \ref{fig:dynasd_width} with fixed depth $D = 5$ based on empirical results. As the width increases, we consistently observe an improvement in acceptance rates and speed across different model pairs. The acceptance rate curve converges when $W = 12$, and the speed curve converges in $W = 10$. These findings demonstrate our method's effectiveness in improving acceptance rates and speed with relatively small tree widths. In addition, we notice there is a noticeable drop in speed when $W = 21$. This speed decline is due to the increased computational overhead associated with processing a more significant number of candidate tokens in parallel, which begins to outweigh the benefits of speculative decoding. \par
\subsubsection{Acceptance Rate}
We run acceptance rate $\alpha$ test across all three datasets with two different temperatures, shown in Table \ref{table:dynasd_acc_rate}. Table \ref{table:acc_rate_comparison} shows our method consistently demonstrating higher acceptance rate $\alpha$ across various datasets compared to baseline SD and MCSD. This indicates that our method's dynamic depth adjustment and target model initialization significantly enhance its ability to generate sequences more aligned with the target model’s expectations. This leads to a higher overall acceptance rate, even under the same draft generation constraints.\par
\subsection{Decision Model}
\label{app:decision_model}
\begin{figure}[h!]
    \centering
    \includegraphics[width=0.7\linewidth]{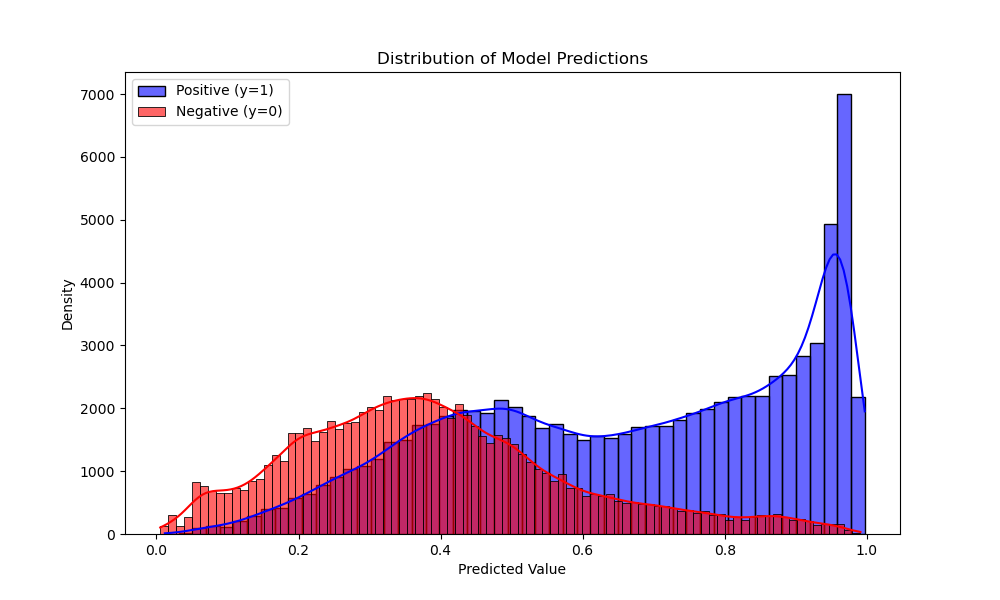}
    \caption{This figure demonstrates that, when using hidden states as input with labels of 0 or 1, the three-layer neural network struggles to effectively distinguish between negative and positive examples. As a result, a significant area of overlap is observed in the predictions. }
    \label{fig:hard_label}
\end{figure}
\begin{figure}[h!]
    \centering
    \includegraphics[width=0.9\linewidth]{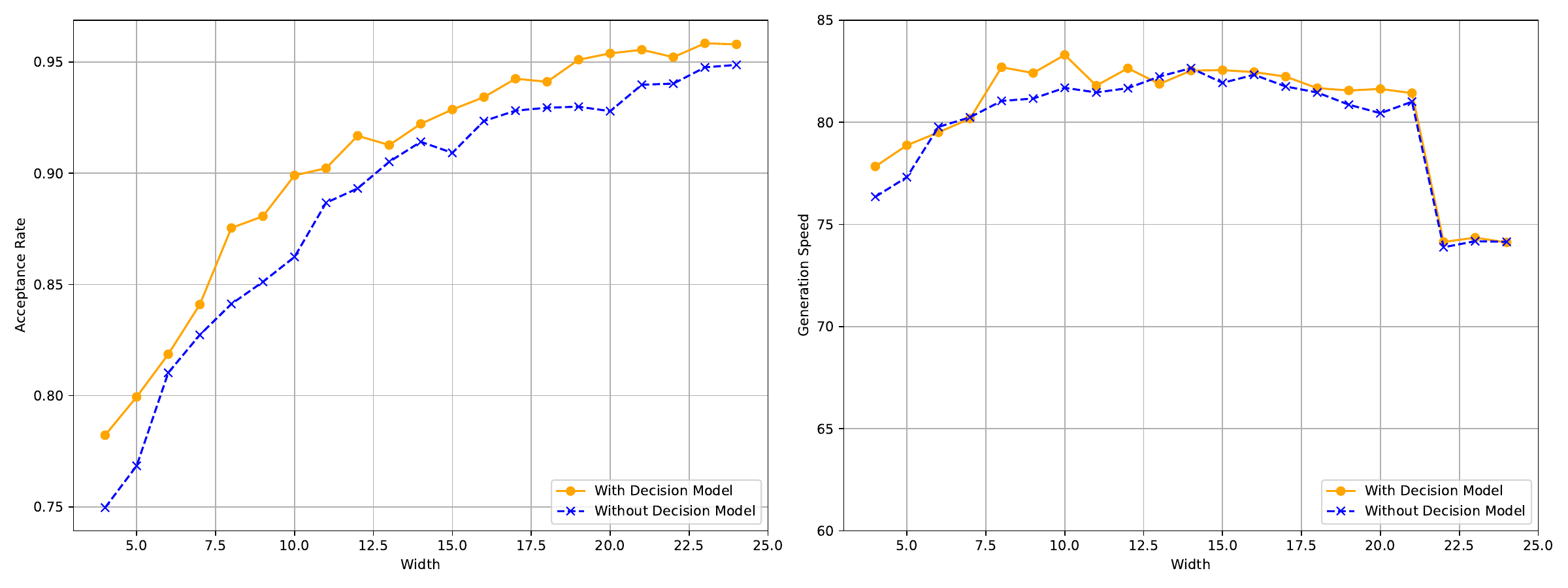}
    \caption{The left figure compares the acceptance rate, whether or not the decision is used, and the right figure compares its speed.}
    \label{fig:decision_model}
\end{figure}
To show the effect of the decision model in our framework, we compare the inference speeds under two scenarios, using or not using the decision model with different parameter size draft models. We still use MT-bench as input prompts, and all other settings remain identical. As a result, in Figure~\ref{fig:decision_model}, both large and small draft models return higher speeds without using the decision model most of the time. The decision model only improves the speed slightly when the width is low. Nonetheless, in either case, the speed improvement of the decision model is minimal. In addition, it proves that our method's speedup is mainly due to the target model's initialized multi-candidate token tree and topology-aware causal mask rather than the decision model.
\clearpage
\bibliographystyle{abbrv}
\bibliography{ref}
\end{document}